\documentclass[runningheads]{llncs}
\usepackage[T1]{fontenc}
\usepackage{booktabs}
\usepackage{siunitx}
\usepackage{makecell}
\usepackage{graphicx}
\usepackage{amsmath}
\usepackage{amssymb}
\usepackage{multirow}
\usepackage{subcaption}
\usepackage{url}
\usepackage{hyperref}
\usepackage{cleveref}

\everypar\expandafter{\the\everypar\loosness=-1}
\begin{document}
\title{PONTE: Personalized Orchestration for Natural Language Trustworthy Explanations}
\titlerunning{PONTE}
%
\author{
Vittoria Vineis\thanks{Equal Contribution. Corresponding Author:  \email{vineis@diag.uniroma1.it}. \\ Preprint. Accepted for publication at the 4th World Conference on Explainable Artificial Intelligence (XAI 2026). Final version to appear in the conference proceedings.
}\inst{1} \and
Matteo Silvestri\textsuperscript{$\star$}\inst{1} \and
Lorenzo Antonelli\textsuperscript{$\star$}\inst{1,2} \and
Filippo Betello\inst{1} \and
Gabriele Tolomei\inst{1,3}
}
\authorrunning{Vineis, Silvestri, Antonelli et al.}
\institute{Sapienza University of Rome \\ 
\and
Syllotips
\and
TellmewAI s.r.l.}
\maketitle

\begin{abstract}
Explainable Artificial Intelligence (XAI) seeks to enhance the transparency and accountability of machine learning systems, yet most methods follow a one-size-fits-all paradigm that neglects user differences in expertise, goals, and cognitive needs. Although Large Language Models can translate technical explanations into natural language, they introduce challenges related to faithfulness and hallucinations. To address these challenges, we present \textbf{PONTE} (\textbf{P}ersonalized \textbf{O}rchestration for \textbf{N}atural language \textbf{T}rustworthy \textbf{E}xplanations), a human-in-the-loop framework for adaptive and reliable XAI narratives. PONTE models personalization as a closed-loop validation and adaptation process rather than prompt engineering. It combines: (i) a low-dimensional preference model capturing stylistic requirements; (ii) a preference-conditioned generator grounded in structured XAI artifacts; and (iii) verification modules enforcing numerical faithfulness, informational completeness, and stylistic alignment, optionally supported by retrieval-grounded argumentation. User feedback iteratively updates the preference state, enabling quick personalization.
Automatic and human evaluations across healthcare and finance domains show that the verification–refinement loop substantially improves completeness and stylistic alignment over validation-free generation. Human studies further confirm strong agreement between intended preference vectors and perceived style, robustness to generation stochasticity, and consistently positive quality assessments.
\end{abstract}


\section{Introduction}
Machine learning (ML) systems are increasingly embedded in decision-making contexts, including online recommendation platforms, financial services, and healthcare \cite{chiusi2020automating}. As these systems increasingly affect individuals and society, concerns about their transparency, fairness, and accountability have grown \cite{dwivedi2023explainable}. In parallel, the European Union has embedded transparency requirements within an evolving regulatory landscape \cite{nannini2024operationalizing}. The GDPR \cite{voigt2017eu} grants individuals the right to “meaningful information about the logic involved”, a principle reinforced by the EU Artificial Intelligence Act \cite{act2024eu}, which mandates transparency and human oversight for high-risk systems. To meet this demand for transparency and accountability, the field of Explainable Artificial Intelligence (XAI) has developed a wide range of techniques aimed at revealing how complex models arrive at their decisions \cite{speith2022review}. However, while these methods can provide valuable technical insights, their outputs are often difficult for non-experts to interpret, limiting their practical usefulness in real-world settings \cite{das2020opportunities}. Moreover, these approaches often implicitly adopt a one-size-fits-all paradigm: XAI methods such as feature attribution techniques \cite{kamolov2025feature} and counterfactual explanations \cite{guidotti2022counterfactual} are designed to be universally valid and model-faithful, yet largely agnostic to the end user. Consequently, the same explanation artifact is often delivered across users, independent of their expertise, objectives, or decision context.
Recently, Large Language Models (LLMs) have emerged as a promising means to overcome these limitations by translating technical XAI outputs into accessible natural language \cite{cedro2025graphxain,pmlr-v258-giorgi25a,giorgi2025enhancingxainarrativesmultinarrative,martens2025tell}. This convergence has given rise to \emph{XAI Narratives} paradigm, where structured explanation artifacts are rendered into human-centered natural-language explanations \cite{silvestri2025survey}. However, integrating LLMs into XAI introduces substantial challenges, including (i) preserving fidelity to the underlying explainer, (ii) ensuring context-appropriate and effective communication of technical XAI outputs, and (iii) maintaining reliability in the argumentation used to support user understanding. Addressing these challenges requires rethinking the role of the human as an active participant in the design and refinement of explanations, further enabled by the accessibility of natural language over structured technical artifacts. However, research on XAI Narratives is still nascent, and existing approaches rarely address these issues in an integrated and systematic way.\\
 To fill this gap, we propose \textbf{PONTE} (\textbf{P}ersonalized \textbf{O}rchestration for \textbf{N}atural language \textbf{T}rustworthy \textbf{E}xplanations), a human-in-the-loop framework for generating adaptive and faithful XAI Narratives. Rather than optimizing for a ``universal'' user, we frame personalization as a closed-loop interaction where coordinated components ensure reliable explanations and alignment with users' needs and preferences.
Concretely, we make the following contributions:\\
(i) \textit{Closed-loop personalization.} We formalize explanation generation as an iterative human–system process in which a dynamic latent state explicitly parameterizes the narrative realization of explanations, ensuring continuous alignment between model-derived facts and the user's stylistic and cognitive requirements.\\
(ii) \textit{Model- and explainer-agnostic orchestration.} We propose a modular orchestration system that transforms structured local post-hoc XAI artifacts into adaptive natural-language narratives, remaining agnostic to both the predictive model and the XAI technique, and thus being applicable across diverse use cases.\\
(iii) \textit{Fidelity-enforcing validation.} We introduce a deterministic mechanism that ensures alignment between the generated narrative and the source XAI output.\\
(iv) \textit{Retrieval-grounded argumentation.} We integrate a retrieval-augmented generation (RAG) step as a modular safeguard to constrain explanatory claims to certified domain literature, reducing reliance on unconstrained parametric knowledge and mitigating plausibility-masked hallucinations.
    
\section{Background and Prior Work}
While earlier XAI literature focused on the goal of providing objective explanations that transparently and accurately describe the behavior of black-box models \cite{guidotti2018survey}, it is now widely acknowledged that 
``one-size-fits-all'' approaches are insufficient to ensure deep understanding and actionability  \cite{liao2021human}. 
A deep analysis of XAI-related user studies confirms that technical interpretability, in fact, does not always entail user comprehension or trust \cite{rong2023towards}, 
as the effectiveness of an explanation is inherently subjective and tied to the specific ``desiderata'' of different stakeholder classes \cite{langer2021we} and unique use-case scenarios \cite{rebstadt2022towards,kong2024toward}. Expanding on this multidimensionality, Suresh et al. \cite{suresh2021beyond} define a stakeholder’s knowledge and their objectives as two cross-cutting components that determine explainability needs: they categorize stakeholder knowledge into formal, instrumental, and personal domains, while proposing a multi-level typology for objectives, ranging from long-term goals like building trust to specific tasks such as detecting mistakes or assessing prediction reliability. Moving along this direction, Speith et al. \cite{speith2024conceptualizing} call for a shift toward an abilities-based conceptualization of understanding, where explanations are not merely passive descriptions but are specifically tailored to help users recognize, assess, or intervene in system behavior. Despite general agreement that explanations should be contextualized and tailored to user needs and objectives to be effective, the mechanisms for achieving this personalization remain largely experimental and varied. While some studies investigate fine-grained personalization based on individual traits \cite{conati2021toward}, others caution against this ``rabbit hole'' suggesting that the focus should instead remain on broader roles and domain-specific requirements rather than minute user details 
\cite{nimmo2024user}. Across approaches, human-in-the-loop mechanisms remain central. These are often implemented through interactive optimization frameworks, such as multi-objective Bayesian optimization guided by iterative user feedback \cite{chandramouli2023interactive} or comparative explanations that expose trade-offs aligned with user priorities \cite{chakraborty2025comparative}. Nevertheless, most of these methods still rely on conventional XAI pipelines, producing technical artifacts that demand substantial cognitive effort and domain expertise to interpret. To bridge this gap, recent research has explored more expressive and natural communication channels, with LLMs and conversational systems emerging as pivotal tools for bridging this divide.
By translating traditional XAI artifacts into fluent, audience-adapted XAI Narratives \cite{silvestri2025survey,martens2025tell}, these systems can broaden interpretability for non-expert users and facilitate more natural human-AI interaction \cite{manohara2024human}. Not surprisingly, recent studies highlight that such narrative approaches are often preferred by users over standard XAI visualizations \cite{silva2024towards}. The use of LLM for XAI, though, introduce complex challenges at the intersection of traditional XAI objetives related to \emph{human utility}, Natural Language Generation concerns around style and phrasing, and, crucially, the need to preserve \emph{faithfulness} to the underlying explainer outputs \cite{silvestri2025survey}. 

As will be discussed, PONTE provides a unified solution to the key challenges identified so far. Before proceeding, however, it is important to distinguish our approach from the use of LLMs as autonomous explainers (see \cite{bilal2025llms}). As shown by Mayne et al. \cite{mayne2025llms}, LLMs lack access to a model’s internal decision boundary, exhibiting a fundamental epistemic blindness that precludes the generation of reliable XAI outputs. In minimal counterfactual generation, models often over-edit inputs or fail to flip predictions under constraints, indicating poor understanding of the decision boundary. Likewise, strong feature attributions may reflect few-shot cues or dataset familiarity, especially on contaminated benchmarks \cite{silvestri2025evaluatinglatentknowledgepublic}, rather than faithful reasoning. This motivates using the LLM only to translate verifiable XAI outputs, not as an independent explanatory agent.

\section{Proposed Solution}
PONTE is a modular orchestration framework for generating personalized natural-language explanations (hereafter, \textit{narratives}) from structured local post-hoc XAI artifacts. Rather than relying solely on prompt engineering, PONTE enforces explanatory reliability and user alignment through explicit validation and grounding mechanisms embedded in a closed-loop architecture.
At a high level (Figure \ref{fig:schema}), the framework takes as input a prediction produced by a black-box model and the corresponding local explanation artifact generated by an explainer. These structured artifacts are processed by a \textit{narrative generation module} conditioned on an explicit user-based \textit{contextual preference model}. The resulting candidate narrative is then subjected to three complementary control stages: a \emph{faithfulness verifier} module, which deterministically verifies alignment with the underlying explanation artifact; a \emph{retrieval-grounded argumentative module}, which substantiates supporting claims through certified domain-specific literature; and a \emph{style alignment verifier}, which ensures consistency with the user’s specified communication preferences.
The validated narrative is delivered to the user, whose feedback updates the preference model, thereby closing the personalization loop. To ensure broad applicability, our system remains predictive-model-agnostic and XAI-method-agnostic. We assume the predictive model to be sufficiently accurate and the selected XAI method to provide reliable explanation artifacts, as those are essential prerequisites for building a trustworthy and personalized narrative layer. The main components of PONTE are described below.\\

\begin{figure}[t]
    \centering
    \includegraphics[width=\linewidth]{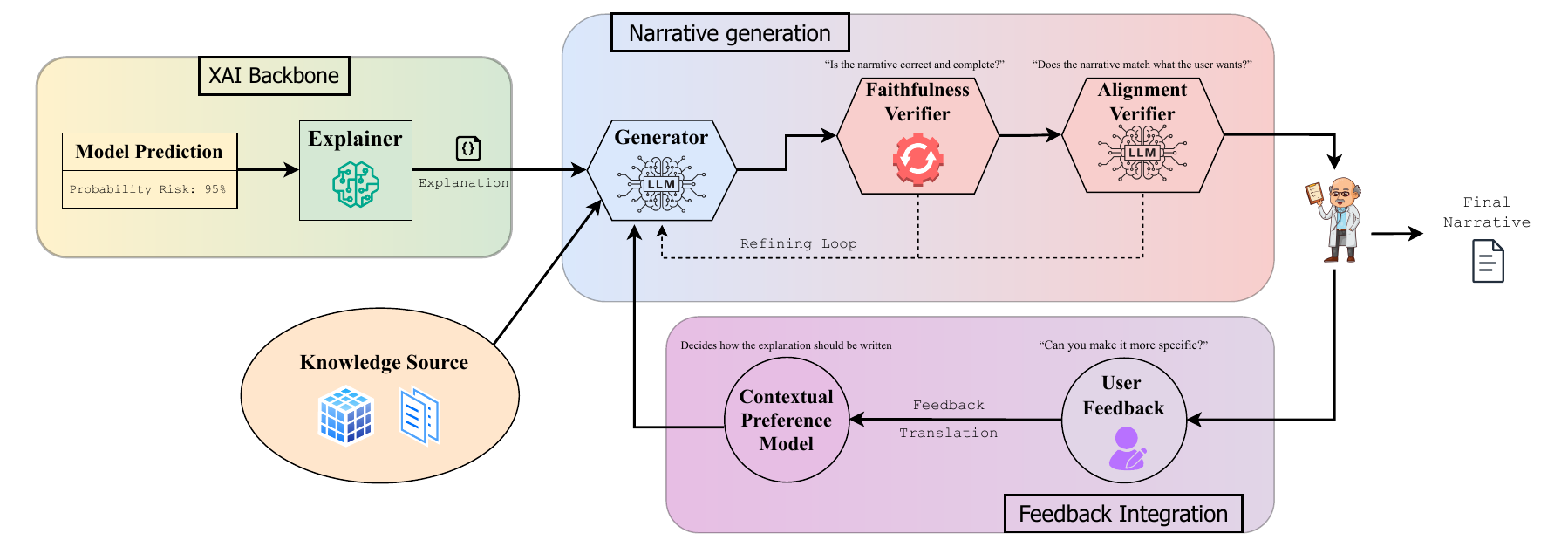}
    \caption{Visual overview of PONTE.}
    \label{fig:schema}
\end{figure}


\noindent \textbf{XAI Backbone.}
The XAI backbone implements the prediction–explanation layer by coupling a predictive model with a local explanation method. Given an input instance, the predictive model produces an output prediction, which is subsequently analyzed by a local explainer to generate structured explanation artifacts, such as feature-attribution scores or counterfactual modifications.
The backbone outputs (i) the input instance, (ii) the corresponding model prediction, and (iii) structured local explanation artifacts. These artifacts constitute the authoritative explanatory substrate for the downstream modules.\\

\noindent \textbf{Contextual Preference Model (CPM).}
This module models personalization as a dynamic preference manifold, providing a low-dimensional latent state that guides how explanations are narrated to meet user-specific needs. User preferences are encoded as a bounded vector $s \in [0,1]^4$, whose components correspond to four interpretable dimensions synthesized from prior literature analysis \cite{silvestri2025survey}: \emph{technicality}, \emph{verbosity}, \emph{depth}, and \emph{actionability}.

\textit{Technicality} controls the level of technical language and data usage, ranging from accessible, plain-language descriptions to quantitatively explicit, domain-specific reporting.  
\textit{Verbosity} regulates narrative flow and length, from telegraphic and minimally connected statements to extended, continuous prose with explicit connective structure.  
\textit{Depth} captures analytical complexity, ranging from isolated feature-level statements to systemic explanations that articulate interactions among multiple factors driving the prediction.  
\textit{Actionability} governs future orientation and prescriptiveness, ranging from diagnostic descriptions of the current decision to actionable guidance framed as concrete steps toward a desired counterfactual outcome. The resulting preference state is exported as a dimension-labeled dictionary, serving both as a conditioning signal for narrative generation and a target reference for downstream style alignment verification.

To mitigate the cold-start problem, the preference vector is initialized from a context-conditioned archetypal persona (e.g., Patient vs. Clinician), providing a structured prior over the style space. This prior is then refined through a closed-loop feedback mechanism, allowing the vector to gradually converge toward user-specific stylistic requirements over a small number of interactions.\\


\noindent \textbf{Narrative Generator.}
This module implements the language realization stage of PONTE, transforming structured XAI artifacts into narratives aligned with user preferences. It generates explanations by supplying the structured explanation payload and the contextual preference vector from the CPM as controlled inputs to a LLM. The structured payload, comprising prediction outputs and local explanation artifacts (e.g., feature importance values, counterfactual changes, and probability variations), is incorporated into the generation context in a controlled, tag-based format to preserve semantic fidelity. When enabled, a retrieval component injects relevant domain knowledge, grounding claims without altering backend, derived explanatory facts.\\

\noindent \textbf{Retrieval-Grounded Argumentation.}
This module substantiates explanatory claims that extend beyond the structured XAI artifact through external, domain-specific knowledge. Rather than relying solely on the LLM’s parametric knowledge, which may mask hallucinations under a plausibility effect \cite{ichmoukhamedov2024good}, it incorporates a RAG mechanism \cite{lewis2020retrieval} over a tailored certified knowledge base tailored to the specific use case (e.g., peer-reviewed medical literature). This constrains argumentative content to externally validated sources, adding an additional layer of epistemic safeguard. Given a candidate narrative and its corresponding explanation artifact, the module formulates targeted retrieval queries and injects semantically relevant evidence into the generation context.\\

\noindent{\textbf{Verifiers.}}
The verifier stage implements a rejection-sampling control layer that evaluates each generated narrative against explicit reliability and alignment constraints. It is composed of two complementary modules.

The \emph{Faithfulness Verifier} enforces both numerical correctness and informational completeness with respect to the underlying explanation artifact. \textit{Correctness} is ensured by deterministically parsing the generated narrative through mandatory tags for quantitative elements (e.g., current values, counterfactual targets, SHAP impacts, and probability shifts). Extracted values are compared against the backend ground-truth explanation under configurable tolerances, turning each explicit numerical claim into a verifiable constraint.
\textit{Completeness} requires that all features modified between the current and counterfactual states are explicitly referenced in the narrative. 

The \textit{Style Alignment Verifier} assesses whether the generated narrative adheres to the stylistic targets defined by the CPM. Style is operationalized as a measurable latent construct along the four rubric-based dimensions. An LLM-based evaluator scores the narrative according to this rubric, and the system computes its deviation from the target preference vector. If the deviation exceeds a predefined tolerance threshold, structured corrective feedback is injected into the orchestration loop to guide refinement.

Constraint violations initiate a controlled rejection–refinement cycle: structured feedback derived from the failing dimensions is incorporated into a corrective prompt, guiding the generation of a revised candidate narrative. The process iterates until all constraints are satisfied or a maximum number of refinement steps is reached.\\

\noindent{\textbf{Feedback Integrator.}}
This module implements the adaptation stage. After the validated narrative is delivered, user feedback, expressed in natural language, is parsed and mapped onto updates of the CPM, through adjustments of the preference vector. These updates modify the control state for subsequent generations, enabling progressive alignment with user-specific communication needs. Through this mechanism, personalization evolves over successive interactions, allowing the system to move beyond the initial archetypical prior toward a refined, user-specific style profile. Once the user confirms that the explanation meets their expectations, the current preference vector is retained as the stable personalized profile for subsequent interactions.

\section{Empirical Validation}
To prove the effectiveness and generalizability of PONTE, we apply it to two representative high-stakes XAI scenarios: (i) a healthcare use case, where a model estimates diabetes risk from clinical attributes \cite{diabetes_dataset}, and (ii) a finance use case, where a model estimates insolvency risk (i.e., loan default probability) from financial data \cite{lendingclub_dataset}. In both settings, we generate local explanations covering feature attributions and counterfactual examples to demonstrate adaptability across different XAI techniques. The experimental setup and evaluation procedure are described in the next subsections.
\subsection{Experimental Setup}
For both scenarios, numerical features are standardized and categorical variables encoded prior to training. The predictive backbone consists of a Multi-Layer Perceptron classifier trained with cross-entropy loss and optimized using Adam on a held-out test split. Local explanations are generated using SHAP \cite{lundberg2017unified} for feature attribution and DiCE \cite{mothilal2020explaining} for counterfactual example generation. Narrative generation and style evaluation are performed using LLM (Kimi-k2.5 \cite{kimiteam2026kimik25visualagentic} and GPT-OSS-20b \cite{gptoss120bgptoss20bmodel}) accessed via Ollama. When enabled, RAG step grounds argumentative content on curated, domain-specific corpora (peer-reviewed medical literature for healthcare; financial and regulatory documentation for finance). Numerical faithfulness is enforced via deterministic value matching under fixed tolerances of 0.05, while stylistic alignment is assessed using an LLM-based rubric scorer with a predefined deviation threshold of $0.2$ and a bounded refinement budget of 10 attempts. User feedback is translated into structured style updates via an LLM-based feedback translator that maps natural-language input into a bounded four-dimensional delta vector \textcolor{black}{$\Delta_t \in [-1,1]^4$}. The CPM is then updated according to  
\textcolor{black}{\(
s_{t+1} = \mathrm{clip}_{[0,1]}\left(s_t + \eta \Delta_t\right),
\)
with fixed step size $\eta = 0.2$}. All hyperparameters in PONTE, including update rates, tolerance thresholds, and refinement budgets, are fully tunable through lightweight configuration files. The full codebase with detailed instructions is publicly available on GitHub\footnote{\url{https://github.com/hercolelab/personalized_xai}}.

\subsection{Evaluation Setup}
We evaluate PONTE with automatic metrics and human-centered protocols, measuring constraint satisfaction (faithfulness, completeness, style alignment) and the perceived usefulness and preference alignment of its narratives

\paragraph{\textbf{Automatic Evaluation.}}
For automatic assessment of style alignment and convergence to user's preferences, we adopt an LLM-as-a-judge paradigm \cite{zhuge2024agent,gu2024survey}. Given a target persona vector and description, an evaluator agent iteratively provides persona-aligned feedback on the generated narrative, simulating user preferences in natural language. At each iteration, the system: (i) generates a verified narrative conditioned on the current preference vector, (ii) obtains persona-aligned feedback from the judge model, (iii) updates the CPM, and (iv) checks convergence based on a predefined alignment tolerance. We define convergence as the CPM preference vector falling within a fixed threshold of the target persona vector. To assess robustness, we conduct batch evaluation on the top-$50$ highest-risk test instances, sampling personas per instance. We quantify the operational cost of alignment via the \emph{Efficiency Score} ($\eta$), defined as the ratio of refinement steps to the Euclidean distance between the baseline preference vector $\mathbf{w}_0 = [0.5, 0.5, 0.5, 0.5]$ and the converged state $\mathbf{w}^*$:
\(
\eta = \frac{T}{\lVert \mathbf{w}^* - \mathbf{w}_0 \rVert_2},
\)
where $T$ denotes the number of feedback iterations.

To assess the contribution of verification and refinement components, we also conduct an ablation study comparing the full framework against a \emph{Single-Pass Baseline}: it uses identical prompts, structured XAI inputs, and the same LLMs, but produces the final narrative in a single generation step, without applying faithfulness and completeness checks, or iterative style-alignment refinement.

\paragraph{\textbf{Human evaluation.}} Human evaluation assesses personalization performance and the perceived quality of the generated narratives. Since correctness and completeness are deterministically enforced by the Faithfulness Verifier, they do not require subjective assessment and are therefore excluded from human evaluation, as they are guaranteed by construction. The reliability of retrieval-augmented argumentation is instead indirectly evaluated through participants’ assessments of clarity and logical consistency, which capture whether externally grounded claims are coherent, well-integrated, and supportive of the explanation.\\
We conduct two complementary human studies (for the forms see this \href{https://drive.google.com/drive/folders/1GTNDS-G49NN16S0L51pjQCAz_mevrStM?usp=drive_link}{folder}):

(i) \textit{Interactive Convergence Simulation (N=3).}  
Three evaluators participate in an interactive alignment experiment, assuming predefined personas. They provide natural-language feedback on generated narratives, which is used to iteratively update the CPM. We compare human-driven refinement trajectories with those by the LLM-as-a-judge, analyzing convergence behavior and consistency between automated and human alignment signals.

(ii) \textit{Survey-based Style and Quality Assessment (N=17).} 
A separate group of evaluators completes a structured survey across four rounds spanning the two use-case scenarios. In each round, participants adopt a predefined persona (Patient vs.\ Clinician; Loan Applicant vs.\ Bank officer), each associated with a target preference profile over the four stylistic dimensions at three ordinal levels (Low/Medium/High; encoded 0/1/2). For each input case, participants evaluate a single narrative realization. To control for generation stochasticity, two independent realizations (V1 vs.\ V2) are produced for the same input and target profile; participants are randomly assigned to evaluate either V1 or V2 (between-subjects assessment). For each given narrative, evaluators provide (i) categorical ratings for the four stylistic dimensions and (ii) Likert-scale assessments of perceived quality, utility and satisfaction.\\
As for the metrics, we quantify stylistic alignment on the encoded ordinal scale using normalized mean absolute error (nMAE) and 
normalized bias (nBias). Let $y_i$ be the human rating and $g$ the corresponding generation level, we define:
\[
\text{nMAE} = \frac{1}{2N} \sum_{i=1}^{N} |y_i - g|,
\qquad
\text{nBias} = \frac{1}{2N} \sum_{i=1}^{N} (y_i - g), \qquad \text{Align.}=1-\text{nMAE}
\]
where $N$ is the number of ratings and the factor $2$ corresponds to the
maximum possible ordinal deviation. Thus, $\text{nMAE} \in [0,1]$
and $\text{nBias} \in [-1,1]$.  We additionally report Spearman’s $\rho$ to assess
monotonic association.
Differences between V1 and V2 are tested via participant-level permutation in a between-subjects design: V1/V2 labels are shuffled per participant (preserving all ratings) to form a null distribution of alignment differences under exchangeability.

\subsection{Main Results}

\begin{table}[t]
\centering
\footnotesize
\setlength{\tabcolsep}{3pt}
\renewcommand{\arraystretch}{0.9}

\caption{Automatic evaluation for Single-Pass Generation (Baseline) and PONTE. We report pass rates for faithfulness (Faith), completeness (Comp), and style alignment (Align). For \textsc{PONTE}, we additionally report the average number of refinement steps over successful runs (Steps) and the failure rate (Fail), defined as the percentage of runs that do not satisfy all constraints within a refinement budget of 10 iterations.}
\label{tab:ponte-ablation}
\begin{tabular}{ll|ccc|cc c | cc} 
\toprule
\multirow{2}{*}{\centering\textit{Dataset}} &
\multirow{2}{*}{\centering\textit{Model}} &
\multicolumn{3}{c|}{\textit{Baseline}} &
\multicolumn{5}{c}{\textit{\textsc{PONTE}}} \\
\cmidrule(lr){3-5}\cmidrule(lr){6-10}
& &
\textbf{Faith} & \textbf{Comp} & \textbf{Align} &
\textbf{Faith} & \textbf{Comp} & \textbf{Align} &
\multicolumn{1}{|c}{\textbf{Steps}} & \textbf{Fail} \\ 
\midrule
Diabetes & Kimi-k2.5 & 0.99 & 0.80 & 0.96 & 1.00 & 0.99 & 0.99 & 1.18 & 0.01 \\
Diabetes & GPT-OSS   & 0.97 & 0.95 & 0.39 & 1.00 & 0.97 & 0.94 & 1.78 & 0.05 \\
Credit   & Kimi-k2.5 & 0.99 & 0.88 & 0.97 & 1.00 & 0.98 & 0.99 & 1.12 & 0.02 \\
Credit   & GPT-OSS   & 0.96 & 0.91 & 0.86 & 1.00 & 0.98 & 0.99 & 1.82 & 0.02 \\
\end{tabular}
\end{table}

\paragraph{\textbf{Automatic evaluation.}}
Table \ref{tab:ponte-ablation} demonstrates the effectiveness of the feedback-guided iterative process and the added value of the verification and refinement components. For both PONTE and the Single-Pass Baseline, we report pass rates over 100 instances for each validation aspect (faithfulness, completeness, and style alignment), along with the overall success rate requiring all aspects to pass. Across datasets and generator models, PONTE consistently improves constraint satisfaction over verifier-free generation. Although the baseline already achieves high faithfulness (0.96–0.99), PONTE reaches perfect faithfulness (1.00) in all settings. The largest gains appear in completeness and style alignment. On Diabetes with Kimi, completeness increases from 0.80 to 0.99. On Diabetes with GPT-OSS, style alignment rises from 0.39 to 0.94, converting a major failure mode into strong alignment. Notably, these improvements incur only a modest refinement cost: PONTE requires on average 1.1–1.8 iterations on successful runs (Steps), with very low failure rates (Fail). 

Similarly, across both datasets, the Agent-as-a-Judge assessment demonstrated high reliability and rapid convergence, with success rates of 94\%--98\% and an average of 1.90 iterations to align ($\sigma \approx 0.45$--$0.48$). While iteration counts remained stable, $\eta$ varied across personas, revealing a clear audience-dependent efficiency gap: professional personas (Bank Officers, Clinicians) were more efficient to align ($\eta \approx 1.9$--$2.8$) than layperson personas (Loan Applicants, Patients), who required nearly twice as many iterations per unit distance ($\eta \approx 4.3$--$4.7$). This suggests that, in our setting, professional preferences, though farther from the baseline, lie in more linearly navigable regions of the latent space, whereas layperson preferences require finer iterative adjustments to stabilize.

\begin{table}[t]
\centering
\small
\setlength{\tabcolsep}{5pt}
\caption{Human alignment with target preference vectors.
Dimension rows pool V1 and V2;
$\Delta$ V1-V2 are evaluated via participant-level permutation tests;
p-values appear in parentheses below the corresponding estimates when applicable.}
\begin{tabular}{lcccc}
\toprule
\textbf{Group}
& \textbf{nMAE} 
& \textbf{nBias}
& \textbf{Align.} 
& \textbf{$\rho$} \\
\midrule
\shortstack{V1\\{\scriptsize\phantom{(p=0.000)}}} &
\shortstack{0.221\\{\scriptsize\phantom{(p=0.000)}}} &
\shortstack{0.029\\{\scriptsize\phantom{(p=0.000)}}} &
\shortstack{0.779\\{\scriptsize\phantom{(p=0.000)}}} &
\shortstack{0.841\\{\scriptsize($p<.001$)}} \\
\shortstack{V2\\{\scriptsize\phantom{(p=0.000)}}} &
\shortstack{0.248\\{\scriptsize\phantom{(p=0.000)}}} &
\shortstack{0.057\\{\scriptsize\phantom{(p=0.000)}}} &
\shortstack{0.752\\{\scriptsize\phantom{(p=0.000)}}} &
\shortstack{0.801\\{\scriptsize($p<.001$)}} \\
\midrule
\shortstack{$\Delta$(V1--V2)\\{\scriptsize\phantom{(p=0.000)}}}
 &
\shortstack{-0.028\\{\scriptsize($p=0.579$)}} &
\shortstack{-0.028\\{\scriptsize($p=0.408$)}} &
\shortstack{+0.027\\{\scriptsize --}} &
\shortstack{+0.040\\{\scriptsize($p=0.577$)}} \\
\midrule
Technicality   & 0.143 & 0.107 & 0.857 & -- \\
Verbosity   & 0.217 & 0.085 & 0.783 & -- \\
Depth  & 0.312 & 0.018 & 0.688 & -- \\
Actionability & 0.268 & -0.033 & 0.732 & -- \\
\end{tabular}
\label{tab:style_alignment_combined}
\end{table}

\begin{figure}[t]
    \centering
    
\includegraphics[width=\linewidth]{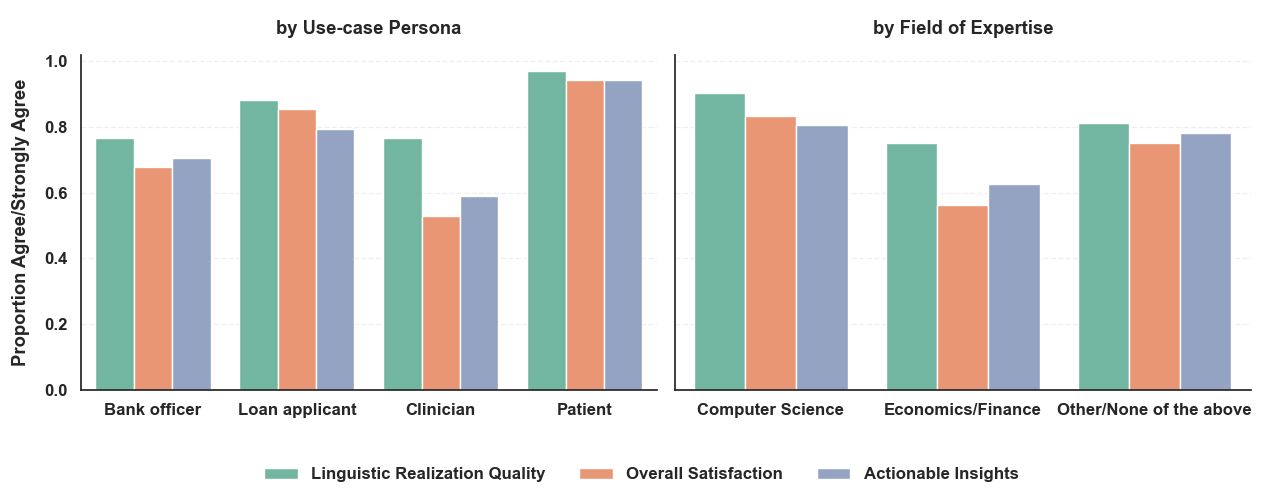}
\caption{Proportion of respondents (N=17) selecting \textit{Agree} or \textit{Strongly Agree} for three evaluation items: \textit{Linguistic Realization Quality} ("The narrative is clear, logically structured, and easy to understand"), \textit{Actionable Insights} ("The narrative helps me understand which changes would be necessary to obtain a different outcome") and \textit{Overall Satisfaction} ("The narrative is satisfactory and appropriate for understanding the model’s decision'"), disaggregated by use-case persona and personal field of expertise.}
    \label{fig:schema}
\end{figure}

\paragraph{\textbf{Human evaluation.}}The convergence behavior observed in the automatic evaluation is confirmed by the human simulations. The average number of iterations among successful runs is 2.33 on Diabetes and 2.0 on LendingClub. With a maximum budget of 10 iterations, the success rates are 75\% and 100\%, respectively. Although humans require slightly more iterations than the agent evaluator, the overall numbers remain low, indicating quick aligment to true user's preferences. This small gap is plausibly explained by the fact that LLM agents are optimized for instruction-following, whereas human participants must first internalize the task and exhibit greater variability in their assessments. Notably, failures occur in the initial rounds of the simulation, suggesting that early task familiarization, rather than instability of the refinement loop, drives non-convergence with the 10-iterations budget.

Table~\ref{tab:style_alignment_combined} reports the alignment between human stylistic assessments and the target preference vectors used by PONTE to generate the narratives. Overall, human evaluations demonstrate strong agreement with the intended stylistic profiles, achieving high alignment scores ($\approx 0.75$--$0.78$). This is further supported by strong monotonic associations (Spearman’s $\rho > 0.80$, $p < .001$), indicating that human-perceived style closely tracks the ordinal structure of the target vectors as instantiated by PONTE in the specific narrative realizations.\\
Moreover, although minor differences are observed between V1 and V2, these differences are not statistically significant (all $p > .40$) under participant-level permutation tests. This indicates that independently sampled realizations (V1 vs.\ V2) of the same input--preference pair yield comparable perceived stylistic profiles. Taken together, these findings suggest that PONTE effectively operationalizes target preference vectors in a way that aligns with human perception, and stylistic alignment is robust to generation stochasticity.
At the dimension level, results reveal heterogeneity across stylistic attributes: technicality shows the highest alignment with the target (Align. = 0.857), followed by verbosity (0.783), whereas depth (0.688) and actionability (0.732) exhibit larger deviations. This pattern suggests that the former dimensions correspond more directly to observable structural features of the text, while depth and actionability reflect higher-order conceptual objectives. As such, the latter likely involve greater semantic complexity and require users to engage in a more subjective interpretation of the argumentative intent rather than relying on surface-level linguistic cues. Across all dimensions, however, normalized bias (nBias) remains small in magnitude, indicating limited systematic over- or underestimation relative to the intended ordinal levels.

Figure \ref{fig:schema} reports the proportion of respondents positively evaluating narrative quality, utility, and overall satisfaction. Endorsement is consistently high across personas, with particularly strong responses for the Patient and Loan Applicant roles, suggesting that the narratives are especially accessible and well-received for non-expert perspectives. When disaggregated by field of expertise, respondents with a Computer Science background (n=9) report systematically higher evaluations across all dimensions. Participants from Economics/Finance field (n=4) exhibit comparatively lower endorsement, particularly for actionable insights and satisfaction, with other expertise (n=4) fall between these groups. Importantly, all subgroup averages remain well above the agreement threshold, indicating broadly positive reception across user profiles.

\section{Conclusion and Future Work}
We presented \textsc{PONTE}, a human-in-the-loop orchestration framework for generating personalized, trustworthy XAI narratives from local explanation artifacts. By framing personalization as a closed-loop validation and adaptation problem with an explicit low-dimensional preference state, PONTE moves beyond prompt engineering and couples preference-conditioned generation with verification modules that enforce numerical faithfulness, informational completeness, and stylistic alignment, optionally grounding supporting argumentation via retrieval over curated domain sources. Empirically, our ablations show that the verifier-refinement loop substantially improves completeness and style alignment over single-pass generation while maintaining near-perfect faithfulness with few refinement steps. Human evaluation further indicates strong agreement between intended preference vectors and perceived style, robustness to generation stochasticity, and broadly positive assessments of clarity, actionability, and satisfaction across use cases and evaluators' backgrounds. Future work may strengthen guarantees for retrieval quality and source attribution and scale to larger user studies and downstream measures of behavioral impact.

\bibliographystyle{splncs04}
\bibliography{biblio}
\end{document}